\documentclass{article}
\usepackage[a4paper, margin={1in}]{geometry}

\usepackage[utf8]{inputenc} % allow utf-8 input
\usepackage[T1]{fontenc}    % use 8-bit T1 fonts

\usepackage{url}            % simple URL typesetting
\usepackage{booktabs}       % professional-quality tables
\usepackage{nicefrac}       % compact symbols for 1/2, etc.
\usepackage{microtype}      % microtypography
\usepackage{lipsum}
\usepackage{fancyhdr}       % header
\usepackage{graphicx}       % graphics

\usepackage[font=small,labelfont=bf]{caption}

\usepackage[bookmarks=false]{hyperref}
    \hypersetup{colorlinks,
      linkcolor=blue,
      citecolor=blue,
      urlcolor=blue}
      
\usepackage[backend=biber, style=numeric]{biblatex}
\usepackage{amsmath}
\usepackage{amssymb}
\usepackage{amsthm}
\usepackage[ruled,vlined]{algorithm2e}

\newcommand{\mat}[1]{\boldsymbol{#1}}
\newcommand{\norm}[1]{\left\lVert\mat{#1}\right\rVert}
\newcommand{\normtilde}[1]{\left\lVert\tilde{\mat{#1}}\right\rVert}
\newcommand{\dotprod}[2]{\mat{#1}^{\top} \mat{#2}}
\newcommand{\dotprodtilde}[2]{\tilde{\mat{#1}}^{\top} \tilde{\mat{#2}}}

\newcommand{\bigO}{\mathcal{O}}
\usepackage[figuresright]{rotating}
\newtheorem{theorem}{Theorem}[section]

\title{Local Random Feature Approximations of the Gaussian Kernel}

\author{
  Jonas Wacker, Maurizio Filippone \\
  Data Science Department, EURECOM, France \\
  \texttt{firstname.lastname@eurecom.fr}
}

\begin{document}

\maketitle

\begin{abstract}
%% Text of abstract
A fundamental drawback of kernel-based statistical models is their limited scalability to large data sets, which requires resorting to approximations.
In this work, we focus on the popular Gaussian kernel and on techniques to linearize kernel-based models by means of random feature approximations.
In particular, we do so by studying a less explored random feature approximation based on Maclaurin expansions and polynomial sketches.
We show that such approaches yield poor results when modelling high-frequency data, and we propose a novel localization scheme that improves kernel approximations and downstream performance significantly in this regime.
We demonstrate these gains on a number of experiments involving the application of Gaussian process regression to synthetic and real-world data of different data sizes and dimensions.
\end{abstract}

%\keywords{Random Features; Gaussian Kernel; Polynomial Kernel; Gaussian Processes}

%\enlargethispage{-7mm}
\section{Introduction}
\label{intro}

% TODO: put OPU paper somewhere

Positive definite kernels are used to model nonlinear phenomena in a theoretically principled way. They have been extensively studied for kernel methods \cite{Schoelkopf2001} as well as for Gaussian processes (GPs) \cite{Rasmussen2006}, where they achieve competitive empirical performance \cite{Rudi2017}. For methods such as Gaussian process regression (GPR), we can obtain closed form predictions by solving linear systems, which is a substantial advantage over deep learning approaches that require iterative solvers and convergence verification.
However, % A well-known drawback of statistical models using kernels is that their 
a naive application relies on algebraic operations on the kernel matrix (or Gram matrix) that consists of pairwise kernel evaluations of the training data. Constructing this matrix therefore requires $\bigO(N^2)$ computations and memory, where $N$ is the number of training points, which obstructs the application of such models when the number of training points is large, e.g., $N > 10 \ 000$.

Hence, considerable effort has been dedicated
% numerous approaches have been proposed 
to improving the scalability of kernel methods and GPs, in particular \cite{williams2001using, Rahimi2007, titsias2009variational, hensman2017variational}. All these methods turn the quadratic dependency on $N$ into a linear or even sub-linear one when using mini-batching, which allows them to scale to millions of data points. A popular line of research uses so-called \textit{random feature (RF)} approximations, which were originally introduced as \textit{random Fourier features} for shift-invariant kernels \cite{Rahimi2007} and were later extended to other classes of kernels such as dot product kernels \cite{Kar2012}.

%A particularly interesting kernel is the popular Gaussian kernel. 
Random feature approximations of the Gaussian kernel, which represents the focus of this work, are well-studied in the literature and are usually based on random Fourier features (c.f. \cite{Liu2020a} for a recent review). However, \cite{Cotter2011} have shown that the Gaussian kernel can also be formulated as a weighted sum of polynomial kernels allowing for a (non-random) Taylor series approximation of the exponential using explicit polynomial basis functions. \cite{wacker2022a} have further shown that using random feature approximations of these polynomial basis functions can make such approaches competitive with random Fourier features provided that the data is scaled appropriately.

In this work, we show that the approach in \cite{wacker2022a} fails for GPR when modelling high-frequency data for which the Gaussian kernel is parameterized by a short length scale. We propose a localized modification of the GPR predictor used in \cite{wacker2022a} that cures this pathology and show that our predictor is competitive and sometimes superior to random Fourier features for a given dimension of the feature map. We evaluate our novel predictor empirically on highly nonlinear synthetic and real-world data (typically modelled using short length scales), and show that it yields state-of-the-art performance regardless of the input dimension of the data. We made our code publicly available: \url{https://github.com/joneswack/dp-rfs}.

Our work is structured as follows. We cover GPR as well as its approximation through a Taylor series approximation with random features in Section~\ref{sec:background}. Our theoretical contributions are made in Section~\ref{sec:methodology}, where we identify and propose a cure for a pathology of such Taylor series approximations.
The empirical evaluation is reported in Section~\ref{sec:experiments}.

\section{Background on Gaussian process regression with Gaussian kernel approximations}
\label{sec:background}

%We begin by introducing Gaussian process regression (GPR), which is a nonparametric Bayesian approach that allows us to model nonlinear data throughout the experiments of this work. We discuss its scalability problems and the use of approximate kernels after.

\subsection{Gaussian process regression (GPR)}
% also explain parametric feature GPs here here
Suppose a training data set $\mathcal{D} := \{(\mat{x}_i, y_i)\}_{i=1}^N$ with $(\mat{x}_i, y_i) \in \mathbb{R}^d \times \mathbb{R}$ that we summarize in matrix notation as $\mat{X} := (\mat{x}_1, \dots, \mat{x}_N)^{\top} \in \mathbb{R}^{N \times d}$ and $\mat{y} := (y_1, \dots, y_N)^{\top} \in \mathbb{R}^N$. We assume that $\mat{y}$ has been generated from $\mat{X}$ by an unknown latent function $f: \mathbb{R}^d \rightarrow \mathbb{R}$ that has been corrupted by independent Gaussian noise, i.e., $y_i = f(\mat{x}_i) + \epsilon_i$ with $\epsilon_i \sim \mathcal{N}(0, \sigma^2_{\rm noise})$ and $\sigma^2_{\rm noise} > 0$.

In GPR \cite[Chapter 2]{Rasmussen2006}, the vector of function evaluations $\mat{\rm f} := (f(\mat{x}_1), \dots, f(\mat{x}_N))^{\top} \in \mathbb{R}^N$ is assumed to have a joint Gaussian distribution with mean $\mat{\mu} \in \mathbb{R}^N$ and covariance matrix $\mat{K}_{\mat{\rm ff}} \in \mathbb{R}^{N \times N}$. We follow the standard approach and set $\mat{\mu}$ to zero here although more complex models exist \cite[Chapter 2.7]{Rasmussen2006}. The entries of $\mat{K}_{\mat{\rm ff}}$ correspond to the evaluations of a positive definite kernel function $k: \mathbb{R}^d \times \mathbb{R}^d \rightarrow \mathbb{R}$, i.e., $(\mat{K}_{\mat{\rm ff}})_{i,j} = k(\mat{x}_i, \mat{x}_j)$ that determines the covariance of a pair of function values $f(\mat{x}_i)$ and $f(\mat{x}_j)$.

The task of GPR is to predict the latent function value at a new test input $\mat{x}_* \in \mathbb{R}^d$ given the training set $\mathcal{D}$. The predictive distribution of $f(\mat{x}_*) | \mathcal{D} \in \mathbb{R}$ can be computed in closed form, and it is $\mathcal{N} (\mu_*, \sigma_*^2 )$ with:
\begin{align}
    \label{eqn:gpr-predictive}
    \mu_* :=  \mat{k}_{\mat{\rm f}*}^{\top} ( \mat{K}_{\mat{\rm ff}} + \sigma^2_{\rm noise} \mat{I})^{-1} \mat{y}
    \quad \text{and} \quad
    \sigma_*^2 := k_{**}  - \mat{k}_{\mat{\rm f}*}^{\top} ( \mat{K}_{\mat{\rm ff}} + \sigma^2_{\rm noise} \mat{I})^{-1} \mat{k}_{\mat{\rm f}*},
\end{align}
where $\mat{k}_{\mat{\rm f}*} = (k(\mat{x}_1, \mat{x}_*), \dots, k(\mat{x}_N, \mat{x}_*))^{\top} \in \mathbb{R}^N$, $k_{**} = k(\mat{x}_*, \mat{x}_*)$ and $\mat{I} \in \mathbb{R}^{N \times N}$ is the identity matrix.
GPR is an attractive modelling choice as it provides uncertainty estimates through the predictive variance $\sigma^2_*$ next to the actual prediction $\mu_*$. At the same time, the hyperparameters of the kernel function $k$ can be obtained through a gradient-based optimization of the \textit{log marginal likelihood} \cite[Chapter 5.4]{Rasmussen2006} avoiding time-consuming cross-validation. 

However, computing the GPR predictor can be expensive in practice. The computational bottleneck is to solve the linear systems in Eq.~(\ref{eqn:gpr-predictive}), which costs $\bigO(N^3)$ time. Even storing the matrix $\mat{K}_{\mat{\rm ff}}$ requires $\bigO(N^2)$ memory and becomes infeasible in practice when $N$ is large, typically greater than $10\ 000$, and approximations become necessary.
%Therefore, the following alternative formulation is often preferred.

\paragraph{Explicit feature space formulation}
If there exists a finite-dimensional feature map $\Phi: \mathbb{R}^d \rightarrow \mathbb{R}^D$ such that $k(\mat{x}, \mat{y}) = \Phi(\mat{x})^{\top} \Phi(\mat{y})$, it can be shown \cite[Chapter 2]{Rasmussen2006} that Eq.~(\ref{eqn:gpr-predictive}) can be reformulated as
\begin{align}
    \mu_* := \Phi(\mat{x}_*)^{\top} \mat{A}^{-1} \Phi( \mat{X})^{\top} \mat{y} / \sigma^2_{\rm noise}
    \quad \text{and} \quad
    \sigma_*^2 := \Phi(\mat{x}_*)^{\top} \mat{A}^{-1} \Phi(\mat{x}_*),
    \ \text{with} \
    \mat{A} := \Phi( \mat{X})^{\top} \Phi(\mat{X}) / \sigma^2_{\rm noise} + \mat{I},
    \label{eqn:gpr-predictive-feature}
\end{align}
where $\Phi(\mat{X}) = (\Phi(\mat{x}_1), \dots, \Phi(\mat{x}_N))^{\top} \in \mathbb{R}^{N \times D}$. The feature space representation (\ref{eqn:gpr-predictive-feature}) changes the computational cost to $\bigO(ND^2)$ and thus improves the scaling of GPR drastically if $D \ll N$. Unfortunately, exact feature maps can be infinite dimensional and this holds in particular for the Gaussian kernel that we study in this work. However, there exist finite dimensional feature maps that yield an \textit{approximate} Gaussian kernel and we discuss them next.

\subsection{Truncated Maclaurin approximation of the Gaussian kernel}
\label{sec:truncated-ml}
% start directly with Maclaurin method and only add RFF if space permits

The Gaussian kernel for two inputs $\mat{x}, \mat{y} \in \mathbb{R}^d$ ($\mat{y}$ is different from the labels in Eq.~(\ref{eqn:gpr-predictive}) here) is defined as $k(\mat{x}, \mat{y}) = \sigma^2 \exp (- \|\mat{x}-\mat{y}\|^2 / (2l^2))$ with its parameters being the length scale $l > 0$ and kernel variance $\sigma^2 > 0$. We rewrite this kernel as a weighted sum of \textit{polynomial kernels} $(\dotprod{x}{y} / l^2)^n$ for $n \in \mathbb{N}$ by using $\|\mat{x} - \mat{y} \|^2 = \norm{x}^2 + \norm{y}^2 - 2 (\dotprod{x}{y})$:
\begin{align}
    \label{eqn:gauss-kernel}
    %\exp \left( - \frac{\|\mat{x}-\mat{y}\|^2}{2l^2} \right)
    k(\mat{x}, \mat{y})
    = \sigma^2 \exp \left( - \frac{\norm{x}^2 + \norm{y}^2}{2 l^2} \right) \exp \left( \frac{\dotprod{x}{y}}{l^2} \right)
    = \sigma^2 \exp \left( - \frac{\norm{x}^2 + \norm{y}^2}{2 l^2} \right) \sum_{n=0}^\infty \frac{1}{n!} \left(\frac{\dotprod{x}{y}}{l^2}\right)^n,
\end{align}
where the second equality of Eq.~(\ref{eqn:gauss-kernel}) follows from the Maclaurin series (Taylor series around zero) of the exponential function. In the following, we obtain a \textit{finite-dimensional} feature map for an approximate Gaussian kernel through \textit{explicit} feature maps for polynomial kernels.

\paragraph{Explicit feature map of the polynomial kernel}
Let $\mat{a} \otimes \mat{b} = \mathrm{vec}(\mat{a} \mat{b}^{\top}) \in \mathbb{R}^{d^2}$ be the vectorized outer product of two vectors $\mat{a}, \mat{b} \in \mathbb{R}^d$. We further define $\mat{a}^{(n)} := \mat{a} \otimes \cdots \otimes \mat{a} \in \mathbb{R}^{d^n}$ to be the result of applying this operation a total number of $(n-1)$ times to a vector with itself. To simplify the notation, we absorb the length scale in the input data, i.e., we define the inputs $\tilde{\mat{x}} := \mat{x} / l$ and $\tilde{\mat{y}} := \mat{y} / l$. Then the polynomial kernel in Eq.~(\ref{eqn:gauss-kernel}) can be written as $(\dotprod{x}{y} / l^2)^n = (\dotprodtilde{x}{y})^n = (\tilde{\mat{x}}^{(n)})^{\top} (\tilde{\mat{y}}^{(n)})$ \cite[Proposition 2.1]{Schoelkopf2001}, where $\tilde{\mat{x}}^{(n)}$ and $\tilde{\mat{y}}^{(n)}$ are its explicit feature maps. We can now use the explicit feature maps for polynomial kernels to obtain an explicit feature map for the Gaussian kernel.

\paragraph{Explicit feature map of the Gaussian kernel}
If we truncate the inifinite Maclaurin series in Eq.~(\ref{eqn:gauss-kernel}) to a finite degree $p \in \mathbb{N}$, we obtain an approximate Gaussian kernel $k_p(\mat{x}, \mat{y}) = \Phi(\mat{x})^{\top} \Phi(\mat{y})$ with an explicit feature map defined as:
\begin{align}
    \label{eqn:explicit-gaussian-features}
    \Phi(\mat{x}) :=
    \sigma \exp \left( - \|\tilde{\mat{x}}\|^2 / 2 \right)
    \left(1, (\tilde{\mat{x}}^{(1)} / \sqrt{1!})^{\top}, \dots, (\tilde{\mat{x}}^{(p)} / \sqrt{p!})^{\top} \right)^{\top} \in \mathbb{R}^{D}
    \quad \text{with} \quad
    D = 1 + d^1 + \dots + d^p.
\end{align}
The approximation error $|k_p(\mat{x}, \mat{y}) - k(\mat{x}, \mat{y})|$ depends on the truncation degree $p$ of the Maclaurin series in Eq.~(\ref{eqn:gauss-kernel}). If $\dotprodtilde{x}{y} \in \mathbb{R}$ is far away from zero, a large $p$ is needed for $\sum_{n=0}^p (\dotprodtilde{x}{y})^n / n!$ to be an accurate estimate of $\exp(\dotprodtilde{x}{y})$. As the dimension $D$ of $\Phi(\mat{x})$ scales as $\bigO(d^p)$, it becomes infeasible to construct such feature maps in practice when $d$ or $p$ are large. Thus, this approach was considered less efficient than random Fourier features \cite{Rahimi2007} with respect to $D$ in \cite{Cotter2011}.

In the following, we substitute the explicit feature maps $\tilde{\mat{x}}^{(1)}, \dots, \tilde{\mat{x}}^{(p)}$ in Eq. (\ref{eqn:explicit-gaussian-features}) with low-dimensional \textit{random} feature maps to attain good kernel estimates with reasonable $D$ that become competitive with random Fourier features.

\subsection{Optimized random Maclaurin features for the Gaussian kernel}
\label{sec:optimized-ml}

% Instead of using the explicit feature maps $\{\phi_n\}_{n=1}^p$ in Eq. (\ref{eqn:explicit-gaussian-features}) for the polynomial kernels $\{ (\dotprod{x}{y} / l^2)^n \}_{n=1}^p$ in Eq.~(\ref{eqn:gauss-kernel}), \cite{wacker2022a} suggest to use random features, from now on referred to as \textit{polynomial sketches}. 
We consider randomized approximations of polynomial kernels that have been used in \cite{Kar2012, Hamid2014, wacker2022a, wacker2022b}. Unlike for the ones proposed in \cite{Pham2013, Ahle2020}, there are closed form variance formulas available for the former in the literature \cite{wacker2022a} that allow us to optimize the variances of our kernel approximation in the following. All these feature maps yield random unbiased approximations of the polynomial kernel, i.e., we have $\mathbb{E}[\phi_n(\tilde{\mat{x}})^{\top} \phi_n(\tilde{\mat{y}})] = (\dotprodtilde{x}{y})^n$ for a random feature map $\phi_n: \mathbb{R}^d \rightarrow \mathbb{R}^{D_n}$, where the variance of the approximation decreases with an increasing dimension $D_n$. We provide an introduction to such random feature maps, from now on referred to as \textit{polynomial sketches}, in \ref{app:polynomial-sketches}.

Using the polynomial sketches $\{\phi_n (\tilde{\mat{x}})\}_{n=1}^p$ instead of the explicit feature maps $\{ \tilde{\mat{x}}^{(n)} \}_{n=1}^p$ in $\Phi(\mat{x})$ (\ref{eqn:explicit-gaussian-features}) yields the following \textit{randomized} approximate kernel for a given truncation degree $p$:
\begin{align}
    \label{eqn:ml-rf-kernel}
    \hat{k}_p(\mat{x}, \mat{y})
    := \Phi(\mat{x})^{\top} \Phi(\mat{y})
    = \sigma^2 \exp \left( - \|\tilde{\mat{x}}\|^2 / 2 \right) \exp \left( - \|\tilde{\mat{y}}\|^2 / 2 \right) \left(1 + \sum_{n=1}^p \frac{1}{n!} \phi_n(\tilde{\mat{x}})^{\top} \phi_n(\tilde{\mat{y}})\right)
\end{align}
with $\mathbb{E} [\hat{k}_p(\mat{x}, \mat{y})] = k_p(\mat{x}, \mat{y})$, where the expectation is with respect to the random feature distribution. The dimension of the feature map $\Phi(\mat{x})$ is $D=1 + D_1 + \dots + D_p$, where $\{D_n\}_{n=1}^p$ are the number of random features allocated to the polynomial sketches $\{\Phi_n\}_{n=1}^p$.

While the $\{D_n\}_{n=1}^p$ were chosen randomly in the past \cite{Kar2012}, it was shown in \cite{wacker2022a} that their allocation under a given budget $D-1$ has a significant impact on the quality of the kernel approximation. The authors in \cite{wacker2022a} show that the feature allocation task can be formulated as a discrete resource allocation problem for which the so-called \textit{Incremental Algorithm} \cite[p. 384]{FloudasChristodoulosA.PardalosPanosM.2009} can be applied. Due to space limitations, we refer the reader to \cite[Chapter 5.3 and Algorithm 3]{wacker2022a} that describes the procedure of finding an optimal degree $p^*$ and an optical allocation $(D_1, \dots, D_{p^*})$ using a subsample of the training data. We will use this method from now on whenever the input dimension $d$ of the data is at least two, otherwise no approximation is needed and explicit polynomial feature maps Eq.~(\ref{eqn:explicit-gaussian-features}) can be used.

\section{Localized random Maclaurin features for the Gaussian kernel}
% Our contribution
\label{sec:methodology}

In this section, we develop our main theoretical and methodological contribution. We begin by uncovering a pathology of Maclaurin-based approximations of the Gaussian kernel that appears when $\|\tilde{\mat{x}}\|$ or $\|\tilde{\mat{y}}\|$ become large.

\subsection{Pathology of the Maclaurin method}

We derive the following Theorem in \ref{app:thrm-proof} that characterizes the pathology of Maclaurin-based approximations leading to poor GPR predictions for high-frequency data.

% Magnitude of product
\begin{theorem}[Vanishing Maclaurin approximation of Gaussian kernels]
\label{thrm:norm-decay}
The magnitude of the finite Maclaurin approximation $k_p(\mat{x}, \mat{y}) = \sigma^2 \exp(- (\|\tilde{\mat{x}}\|^2 + \|\tilde{\mat{y}}\|^2) / 2) \sum_{n=0}^p 1/n! (\dotprodtilde{x}{y})^n$ of the Gaussian kernel approaches zero as $\|\tilde{\mat{x}}\|^2 + \|\tilde{\mat{y}}\|^2$ increases. The error $|k(\mat{x}, \mat{y}) - k_p(\mat{x}, \mat{y})|$ between the exact kernel $k$ and its approximation $k_p$ is the largest for parallel $\mat{x}, \mat{y}$ and zero when they are orthogonal.
\end{theorem}

\begin{figure}[t]\vspace*{4pt}
\centerline{\includegraphics[width=0.9\textwidth]{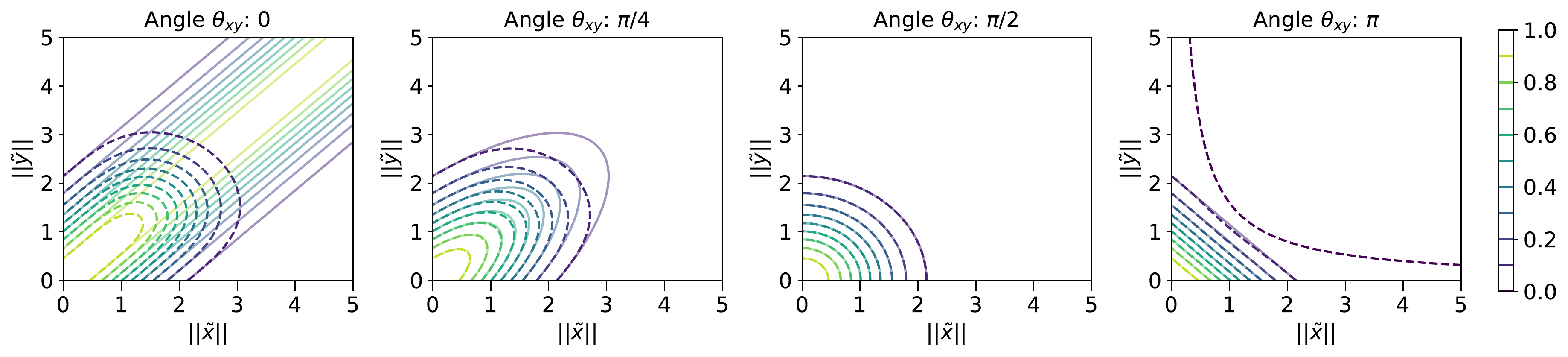}}
\caption{$k_p(\mat{x}, \mat{y})$ (dashed) vs. $k(\mat{x}, \mat{y})$ (transparent and solid) for fixed $p=3$ and $\sigma^2=l=1$, where we use the relationship $\dotprodtilde{x}{y} = \normtilde{x} \normtilde{y} \cos(\theta_{xy})$.}
\label{fig:vanishing-maclaurin}
\end{figure}

We visualize the implications of Theorem~(\ref{thrm:norm-decay}) in Fig.~(\ref{fig:vanishing-maclaurin}), where we compare the approximation $k_p(\mat{x}, \mat{y})$ with the exact Gaussian kernel $k(\mat{x}, \mat{y})$ for $p=3$ over a range of values $\normtilde{x}, \normtilde{y}$ as well as the angle $\theta_{xy}$ between $\mat{x}$ and $\mat{y}$. One can see that $k_p(\mat{x}, \mat{y})$ approaches zero with increasing $\normtilde{x}^2 + \normtilde{y}^2$ regardless of $\theta_{xy}$. This deteriorates the approximation quality, in particular as $\theta_{xy}$ goes to zero. This development accelerates when choosing a shorter length scale $l$.
A consequence of this pathology is that the GPR predictive means and variances in Eq.~(\ref{eqn:gpr-predictive}) collapse to zero for test points $\tilde{\mat{x}^*}$ with large $\|\tilde{\mat{x}^*}\|$ since $k_p(\mat{x}^*, \mat{x})$ goes to zero for any $\mat{x} \in \mathbb{R}^d$ in this case. This effect is shown in the middle plot of Fig.~(\ref{fig:1d-regression}). We will discuss this example in greater detail in Section~\ref{sec:experiments}.

A similar pathology was identified for the \textit{Relevance Vector Machine} \cite{tipping1999} that was solved by adding new basis functions centered on the test points \cite{Rasmussen2005}. In the following, we present a similar strategy by centering our entire approximation around individual test points.

\subsection{Curing the pathology for GP regression}
\label{sec:pathology-cure}

We will now exploit a property of the Gaussian kernel that allows us to cure the aforementioned pathology. The Gaussian kernel is shift-invariant, i.e., $k(\mat{x} + \mat{\delta}, \mat{y} + \mat{\delta}) = k(\mat{x}, \mat{y})$ for any $\mat{\delta} \in \mathbb{R}^d$, because $\|(\mat{x} + \mat{\delta}) - (\mat{y} + \mat{\delta})\| = \|\mat{x} - \mat{y}\|$. Thus, when making a prediction at a test input $\mat{x}_*$, one can subtract $\mat{x}_*$ from all inputs used in Eq.~(\ref{eqn:gpr-predictive}) without changing the result of the prediction. More specifically, the values $\mu_*$ and $\sigma^2_*$ in Eq.~(\ref{eqn:gpr-predictive}) do not change if we substitute $k(\mat{x}, \mat{y})$ by $k(\mat{x} - \mat{x}_*, \mat{y} - \mat{x}_*)$ for the computation of $\mat{k}_{\mat{\rm f*}}, \mat{K}_{\mat{\rm ff}}$ and $k_{**}$.

However, the approximate kernel $\hat{k}_p$ (\ref{eqn:ml-rf-kernel}) is greatly affected by this change. To see this, we define $\hat{k}_p^*(\mat{x}, \mat{y}) := \hat{k}_p(\mat{x} - \mat{x}_*, \mat{y} - \mat{x}_*)$ with:
%
% \begin{align}
%     \label{eqn:kp-star}
%     k_p^*(\mat{x}, \mat{y})
%     := k_p(\mat{x} - \mat{x}_*, \mat{y} - \mat{x}_*)
%     = \sigma^2 \exp \left( - \frac{\|\mat{x}-\mat{x}^*\|^2}{2 l^2} \right) \exp \left( - \frac{\|\mat{y}-\mat{x}^*\|^2}{2 l^2} \right) \sum_{n=0}^p \frac{1}{n!} \left(\frac{(\mat{x}-\mat{x}^*)^{\top} (\mat{y}-\mat{x}^*)}{l^2}\right)^n
% \end{align}
\begin{align}
    \label{eqn:ml-rf-kernel-centered}
    \hat{k}_p^*(\mat{x}, \mat{y})
    = \sigma^2 \exp \left( - \frac{\|\mat{x}-\mat{x}^*\|^2}{2 l^2} \right)
    \exp \left(- \frac{\|\mat{y}-\mat{x}^*\|^2}{2 l^2} \right)
    \sum_{n=1}^p \frac{1}{n!} \Phi_n \left(\frac{\mat{x}-\mat{x}^*}{l} \right)^{\top} \Phi_n \left(\frac{\mat{y}-\mat{x}^*}{l} \right),
\end{align}
where $\mathbb{E}[\hat{k}_p^*(\mat{x}, \mat{y})] = k_p(\mat{x} - \mat{x}^*, \mat{y} - \mat{x}^*) =: k_p^*(\mat{x}, \mat{y})$. As all $\{\Phi_n\}_{n=1}^p$ are sampled independently, we have
\begin{align}
    \label{eqn:variance-kp-star}
    \mathbb{V}[\hat{k}_p^*(\mat{x}, \mat{y})] = \mathbb{V}[\hat{k}_p(\mat{x}-\mat{x}^*, \mat{y}-\mat{x}^*)] \propto \sum_{n=1}^p \left(\frac{1}{n!}\right)^2 \ \mathbb{V}\left[\Phi_n\left(\frac{\mat{x} - \mat{x}^*}{l}\right)^{\top} \Phi_n \left(\frac{\mat{y}  - \mat{x}^*}{l}\right)\right],
\end{align}
where the variance is with respect to the random feature distribution. When setting $\mat{x} = \mat{x}^*$ or $\mat{y} = \mat{x}^*$, the variance terms in Eq.~(\ref{eqn:variance-kp-star}) become 
%$\mathbb{V}[\Phi_n(\mat{0})^{\top} \Phi_n(\mat{y}  - \mat{x}^*)]$ and $\mathbb{V}[\Phi_n(\mat{x}  - \mat{x}^*)^{\top} \Phi_n(\mat{0})]$, respectively. As can be seen from Table~(\ref{tbl:sampling-variances}), this turns the variances to \textit{zero} for any of the polynomial sketches discussed in \ref{app:polynomial-sketches}. $\hat{k}_p^*(\mat{x}^*, \mat{y}), \hat{k}_p^*(\mat{x}, \mat{x}^*)$ and $\hat{k}_p^*(\mat{x}^*, \mat{x}^*)$ thus become \textit{deterministic}.
\textit{zero} for any of the polynomial sketches discussed in \ref{app:polynomial-sketches} as can be seen from Table~(\ref{tbl:sampling-variances}). $\hat{k}_p^*(\mat{x}^*, \mat{y}), \hat{k}_p^*(\mat{x}, \mat{x}^*)$ and $\hat{k}_p^*(\mat{x}^*, \mat{x}^*)$ thus become \textit{deterministic}.

We further have $\hat{k}_p^*(\mat{x}^*, \mat{y}) = \sigma^2 \exp(-\|\mat{x}^* - \mat{y}\|^2 / (2l)^2)$, $\hat{k}_p^*(\mat{x}, \mat{x}^*) = \sigma^2 \exp(-\|\mat{x} - \mat{x}^*\|^2 / (2l)^2)$ and $\hat{k}_p^*(\mat{x}^*, \mat{x}^*) = \sigma^2$ which are all equal to the exact kernel $k$ evaluated at these points. Therefore, $\mat{k}_{\mat{\rm f*}}$ and $k_{**}$ in Eq.~(\ref{eqn:gpr-predictive}) become exact for our Maclaurin approximation. $\mat{K}_{\mat{\rm ff}}$ unfortunately remains affected by the vanishing approximate kernels described by Theorem~(\ref{thrm:norm-decay}) and by non-zero random feature variances.
%, especially when $(\mat{x} - \mat{x}^*)^{\top} (\mat{y} - \mat{x}^*) / l^2$ is large,
%leaving inaccuracies in the approximation.
A crucial advantage of using $\hat{k}_p^*$ instead of $\hat{k}_p$ is that the GP predictive distribution (\ref{eqn:gpr-predictive}) does not collapse to zero anymore as $\|\tilde{\mat{x}^*}\|$ grows. We illustrate this on the following synthetic example data set. As it is one-dimensional, we stick to the deterministic feature map in Eq.~(\ref{eqn:explicit-gaussian-features}) for now.

\paragraph{Approximating the sinc function}

\begin{figure}[t]\vspace*{4pt}
\centerline{\includegraphics[width=1.0 \textwidth]{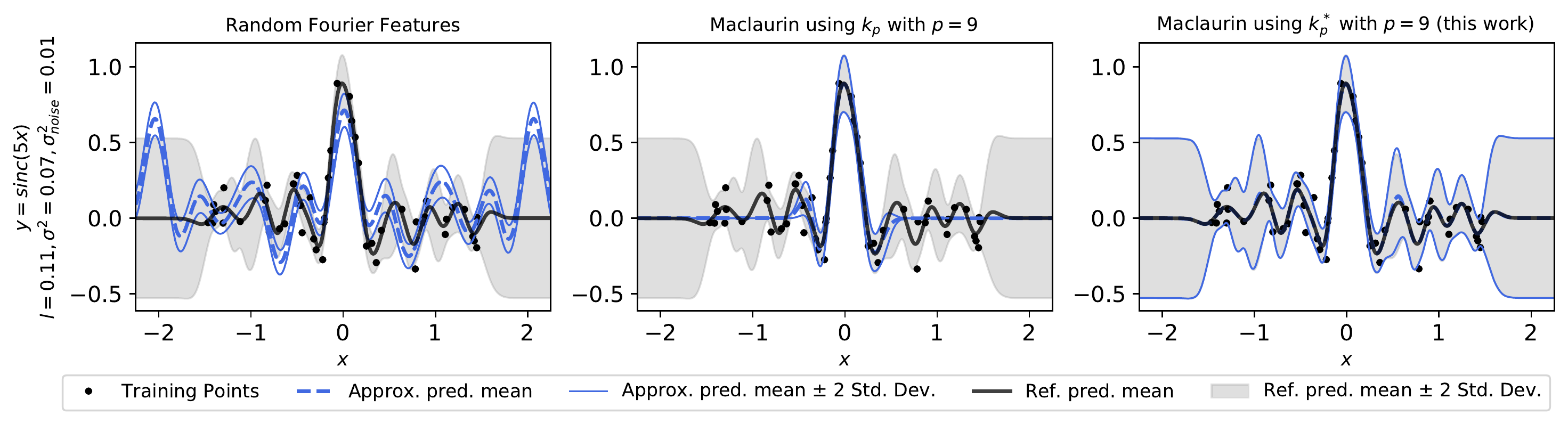}}
\caption{Approximating the predictive distribution of a reference GPR using different approximation methods and $D=10$.}
\label{fig:1d-regression}
\end{figure}

We draw 50 noisy observations $\{y_i\}_{i=1}^{50}$ with $y_i=sinc(5x_i)+\epsilon_i$ and $\epsilon_i \stackrel{i.i.d.}{\sim} \mathcal{N}(0, \sigma^2_{\rm noise} = 0.01)$, where $\{x_i\}_{i=1}^{50}$ are sampled independently and uniformly from the interval $[-1.5, 1.5]$. We then fit a reference GPR on this data set using the Gaussian kernel (\ref{eqn:gauss-kernel}), where the hyperparameters $l$ and $\sigma^2 > 0$ are found through a gradient based optimization of the log marginal likelihood \cite[Chapter 5.4]{Rasmussen2006}.
The length scale found for the reference GPR is $l=0.11$ and is rather short compared to $0.92$, which is the median pairwise Euclidean distance of the training data, a standard heuristic for choosing the length scale without optimization \cite{garreau2017large}. This reflects the frequent oscillations of the function ${\rm sinc}(5x)$.
We show the reference GPR along with three different approximation schemes in Fig.~(\ref{fig:1d-regression}). The baseline Random Fourier features (RFF) \cite{Rahimi2007} (left) struggles to recover the reference GPR for $D=10$ random features. The dimension $D$ of the feature map (\ref{eqn:explicit-gaussian-features}) is equal to $p+1$ for the Maclaurin approximation. So we chose $p=9$ for a fair comparison against RFF.
%$D$ is crucial in determining the cost of computing the GPR predictor in Eq.~(\ref{eqn:gpr-predictive-feature}) and should therefore be kept small.

The Maclaurin approach using $k_p^*$ (right) gives the best approximations while the predictive distribution of the one using $k_p$ (middle) collapses to zero very quickly at points $x^* \in \mathbb{R}$ away from zero. This is because $\|\tilde{x^*}\|$ becomes large very quickly and $k_p$ vanishes when being evaluated at these points.
For values $x^*$ far away from the training data, the GP predictor (\ref{eqn:gpr-predictive}) using $k_p^*$ even recovers the GPR prior distribution as desired, which can be explained as follows. When $x^*$ is far from the training data, $\mat{k}_{\mat{\rm f}*}$ becomes zero. Then $\mu_*=0$ and $\sigma^2_*=k_{**}=\sigma^2$ in Eq.~(\ref{eqn:gpr-predictive}). Since $k_{**}$ and $\mat{k}_{\mat{\rm f}*}$ in Eq.~(\ref{eqn:gpr-predictive}) are accurate when using $k_p^*$ as explained earlier, the convergence to the prior is kept for $k_p^*$.

\subsection{Reducing computational costs through clustering}
\label{sec:clustering}

\begin{algorithm}[t]
\caption{Precomputing Localized Maclaurin Features for the Gaussian Kernel}
    \label{alg:kp-star-algorithm}
    
    \SetAlgoLined
    \textbf{Input:} Hyperparameters $l, \sigma^2, \sigma^2_{\rm noise} > 0$; training data $\{\mat{x}_i\}_{i=1}^N$; test data $\{\mat{x}_i^*\}_{i=1}^{N_*}$; number of features $D \geq 1$;
    
    \tcp*[h]{1) Training} \\
    \If(\tcp*[h]{We use random features}){$d > 1$}{
        Center the training data by subtracting the training mean \;
        Find $(p^*, D_1, \dots, D_{p^*})$, the optimal feature allocation for the random Maclaurin method (see Section~\ref{sec:optimized-ml}) \;
    }
    $C := \{ (\sum_{i=1}^N \mat{x}_i) / N \}$ \tcp*[h]{Initialize centroids with training mean} \;
    \While(\tcp*[h]{Farthest point clustering}){True}{
        Let $\delta_i = \min \{ \|\mat{x}_i - \mat{c}\| \, | \, \mat{c} \in C \}$ for $i=1, \dots, N$ \;
        
        \If(\tcp*[h]{Training Max.-Min.-Distance is below threshold $\theta$})
        {$\max \{\delta_i\}_{i=1}^N < \theta$}{
            break;
        }
        Add $\mat{x_i}$ with $i = \arg \max_i \{\delta_i\}_{i=1}^N$ to $C$ \;
        
        Precompute $\mat{A}_c^{-1} := \left(\Phi( \mat{X} - \mat{c})^{\top} \Phi(\mat{X} - \mat{c}) / \sigma^2_{\rm noise} + \mat{I} \right)^{-1}$ with $\Phi$ defined in Eq.~(\ref{eqn:explicit-gaussian-features}) \;
        
        \tcp*[h]{Use explicit features for $d=1$, else use polynomial sketches (\ref{app:polynomial-sketches})}
    }
    
    \ForAll(\tcp*[h]{2) Inference}){$\{\mat{x}_i^*\}_{i=1}^{N_*}$}{
        Assign $\mat{x}_i^*$ to closest centroid $\mat{c} \in C$ \;
        Compute $\mu_*$ and $\sigma^2_*$ in Eq.~(\ref{eqn:gpr-predictive-feature}) by setting $\mat{A}^{-1} = \mat{A}_c^{-1}$, $\mat{x}^* = \mat{x}_i^* - \mat{c}$ and $\mat{X} = \mat{X} - \mat{c}$ \;
    }
\end{algorithm}

A caveat of using the approximate kernel $\hat{k}_p^*(\mat{x}, \mat{y})$ (\ref{eqn:ml-rf-kernel-centered}) described in Section~\ref{sec:pathology-cure} is that we need to recompute the GPR predictor (\ref{eqn:gpr-predictive}) for every test point $\mat{x}^*$ separately, which becomes expensive when many test points need to be predicted, even when using the featurized version (\ref{eqn:gpr-predictive-feature}) of the predictor. We denote the number of test inputs by $N_*$. A direct computation of Eq.~(\ref{eqn:gpr-predictive-feature}) now costs a total of $\bigO(N_* ND^2)$ for all test points.

The problem is ``embarrassingly parallel'' and the computation of Eq.~(\ref{eqn:gpr-predictive-feature}) for every $\mat{x}^*$ could be easily distributed on a cluster of compute nodes or parallelized using a single GPU, e.g., using JAX \cite{jax2018github}. However, we propose a different approach here that requires no separate training at test time while staying as close as possible to the training time of $\bigO(ND^2)$ as is generally desired for random feature approximations.

Our approach is to cluster
%\footnote{We choose a farthest point clustering for simplicity as it does not require convergence verification and determines the number of clusters using a threshold $\theta > 0$. Any other clustering algorithm can be used instead (even a random selection of training points).} 
the \textit{training} data into $N_{C}$ clusters and to use the centroids of these clusters as \textit{pseudo test inputs}. 
We choose a farthest point clustering for simplicity as it does not require convergence verification and determines the number of clusters using a threshold $\theta > 0$, but any clustering algorithm can be used instead (even a random selection of training points).
As the centroids are known during training, we can pretrain a set of $N_{C}$ predictors using Eq.~(\ref{eqn:gpr-predictive-feature}) and assign a new test point $\mat{x}^*$ to the closest centroid at prediction time. We summarize the complete procedure in Alg.~(\ref{alg:kp-star-algorithm}). The computational cost is now $\bigO(N_C N D^2)$ and is thus much lower than $\bigO(N_* ND^2)$ if $N_C \ll N_*$.

\section{Empirical evaluation on real-world data}
\label{sec:experiments}

% TODO: Explain that we picked high frequency data which is hard to find in high dimensions

In this section, we evaluate our proposed method on real-world data of different dimensions for which we employ the polynomial sketches in \ref{app:polynomial-sketches}. As for the synthetic example in Fig.~(\ref{fig:1d-regression}), we use the Gaussian kernel with hyperparameters $l$ and $\sigma^2 > 0$ that are found through gradient based optimization of the log marginal likelihood of a reference GPR, along with $\sigma^2_{\rm noise}$. 
%As $\sigma^2_{\rm noise}$ is not known from now, we include it in this optimization.

We compare our method against a random Fourier features baseline \cite{Rahimi2007} as well as its structured extension \cite{Yu2016} when the data is sufficiently high dimensional\footnote{Otherwise, the structured random Fourier features induce a large bias.}. We also add the vanilla optimized Maclaurin method (Section \ref{sec:optimized-ml}) to this comparison. It is equivalent to Alg.~(\ref{alg:kp-star-algorithm}) using only a single cluster with its centroid being the training mean. We measure the approximation quality with respect to the reference GPR using the Kullback-Leibler (KL) divergence \cite[Chapter A.5]{Rasmussen2006} between the predictive means and variances in Eq.~(\ref{eqn:gpr-predictive}) and Eq.~(\ref{eqn:gpr-predictive-feature}). We measure downstream regression performance using the root mean squared error (RMSE).

\subsection{UK apartment price data}

We downloaded the monthly property sales data for England and Wales from the HM land registry\footnote{\url{https://www.gov.uk/government/statistical-data-sets/price-paid-data-downloads}}. We filtered for sold apartments for the month of January 2022 leading to a data set with $24 \ 553$ observations. Matching the post codes for each apartment sold with a database of latitudes and longitudes\footnote{\url{https://www.freemaptools.com/download-uk-postcode-lat-lng.htm}} allowed us to obtain a two-dimensional data set (latitude, longitude) that we could regress against the logarithm of the sales prices. We randomly split the data into $10 \ 000$ training points and kept the rest for testing.

\begin{figure}[t]\vspace*{4pt}
\centerline{\includegraphics[width=1.0 \textwidth]{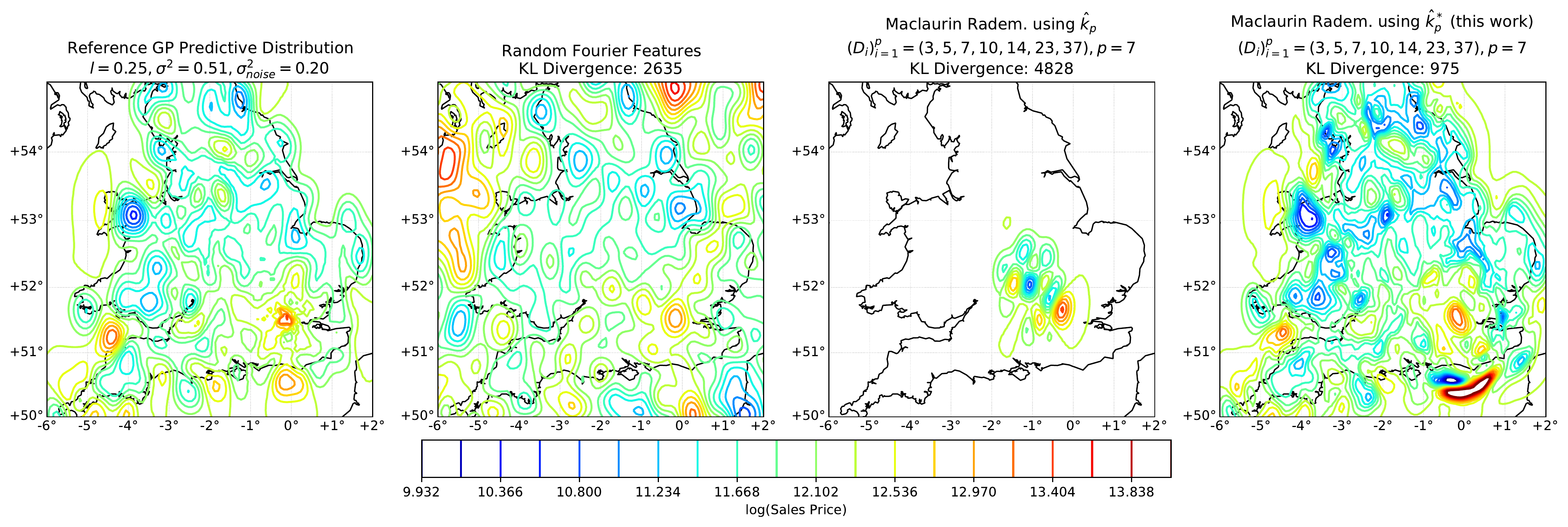}}
\caption{Approximating the predictive mean of a reference GP using different approximation methods and $D=100$. KL Divergences also include predictive variances. The Maclaurin method using $\hat{k}_p$ makes predictions centered around the mean of the training data (close to London).}
\label{fig:uk-house-prices-kp-star}
\end{figure}

In our first experiment, we aim to recover the reference GPR predictive distribution on a regular grid of latitudes (between $+50^{\circ}$ and $+55^{\circ}$) and longitudes (between $-6^{\circ}$ and $+2^{\circ}$) of size $100$ by $100$. Fig.~(\ref{fig:uk-house-prices-kp-star}) shows the results of this experiment. As for the sinc-example in Fig.~(\ref{fig:1d-regression}), random Fourier features struggle to recover the predictive distribution using $D=100$ random features and the vanilla Maclaurin method using $\hat{k}_p$ (\ref{eqn:ml-rf-kernel}) suffers from vanishing kernels due to the short (compared to the scaling of the data) length scale of $l=0.25$. Our proposed kernel $\hat{k}_p^*$ (\ref{eqn:ml-rf-kernel-centered}) improves predictions considerably leading to the lowest KL divergence with respect to the reference GP predictive distribution. It also converges to the prior for test points far from the training data.

In our second experiment, we evaluate the use of Alg.~(\ref{alg:kp-star-algorithm}) to precompute the matrix inversion in Eq.~(\ref{eqn:gpr-predictive-feature}) on a set of pseudo test inputs. This time we report results on the left-out test data instead of a regular grid. The left part of Fig.~(\ref{fig:uk-house-prices-precomputation}) shows these results. We can see that the KL divergence falls off considerably (top plot) as we add more clusters until reaching 57 clusters. From then on the KL divergence remains roughly the same indicating that 57 clusters give a good trade-off between efficiency and performance. In the bottom plot we show a comparison of RMSE values for these 57 clusters, where the Maclaurin method outperforms random Fourier features, in particular for small $D$.

% TODO: mention that results are better than in figure below because of prior convergence

\subsection{UCI data sets: Yacht and kin8nm}

\begin{figure}[t]\vspace*{4pt}
\centerline{\includegraphics[width=0.9\textwidth]{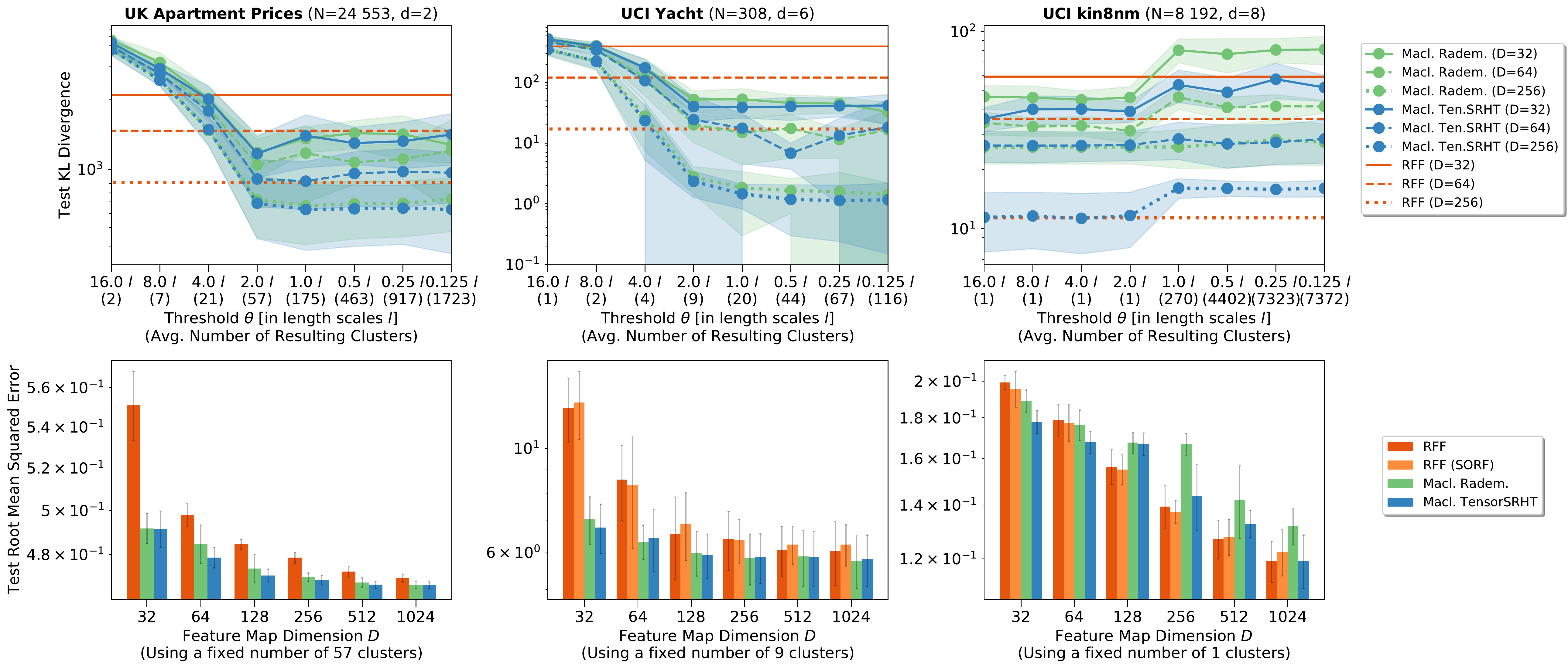}}
\caption{(Top) Test KL Divergence for a given number of clusters obtained from the training data using Alg.~(\ref{alg:kp-star-algorithm}) with threshold $\theta$. (Bottom) RMSE over feature map dimension $D$ for a fixed number of clusters.}
\label{fig:uk-house-prices-precomputation}
\end{figure}

In the following, we repeat the evaluation of Alg.~(\ref{alg:kp-star-algorithm}) for two higher dimensional data sets that are taken from the UCI machine learning repository \cite{Dua:2019} in Fig.~(\ref{fig:uk-house-prices-precomputation}).
We obtain very similar results for the UCI Yacht data set as for the UK apartment price data set. Adding more clusters gives large gains initially but these diminish when setting $\theta > 2.0l$ in Alg.~(\ref{alg:kp-star-algorithm}) determining a good trade-off between performance and computational cost. This time we included structured orthogonal random Fourier features \cite{Yu2016} that are also outperformed by the Maclaurin method.

For the UCI kin8nm data set, results look quite different. Adding more clusters \textit{increases} the KL divergence towards the reference GP predictor, which is why only a single centroid, the mean of the training data, is chosen for the RMSE comparison in the plot below. This corresponds to the vanilla optimized Maclaurin method (Section \ref{sec:optimized-ml}).

We explain this observation as follows. The length scales obtained for the sinc example, the UK apartment price data and for UCI Yacht are short. For the UCI Yacht data set it is 0.32, i.e., much less than 3.28, the median pairwise Euclidean distance of the training data, indicating that the data is fit by a reference GP of high frequency. For kin8nm the length scale is 2.15 compared to 3.93 (median heuristic) indicating a much smoother GP than the ones before.

In this case, the values of $\mat{k}_{\mat{\rm f}*}$ and $k_{**}$ in Eq.~(\ref{eqn:gpr-predictive}) are less affected by the vanishing kernels (Theorem \ref{thrm:norm-decay}) due to a longer length scale. However, the approximation of $\mat{K}_{\mat{\rm ff}}$ in Eq.~(\ref{eqn:gpr-predictive}) is more accurate when using the vanilla Maclaurin approximation (\ref{eqn:ml-rf-kernel}) because the data is centered around the training mean.
This shows that the clusters need to be chosen depending on the smoothness of the target GP. In this work, we have provided a \textit{generalization} of the vanilla Maclaurin method that (with an appropriate choice of clusters) can fit both, high and low-frequency data.

\newpage

\section{Conclusion}

We have identified a major pathology when using Maclaurin-based approximations such as \cite{Cotter2011, wacker2022a} for the Gaussian kernel. We have further presented an extension of the optimized Maclaurin method \cite{wacker2022a} that overcomes this problem and makes it applicable to high-frequency data. The clustering method in Alg.~(\ref{alg:kp-star-algorithm}) seems to have a strong impact on predictive performance. Future work should investigate on optimal clustering schemes that automatically adapt to the frequency of the target function. It would further be interesting to combine the advantages of random Fourier features and polynomial sketches, as both approximations have zero variances in different regimes (equal inputs for RFF and orthogonal inputs for Maclaurin).

%\section*{Acknowledgements}
\paragraph{\bf Acknowledgements}
We thank Sagar Arora for helpful discussions. MF gratefully acknowledges support from the AXA Research Fund and the Agence Nationale de la Recherche (ANR-18-CE46-0002 and ANR-19-P3IA-0002).

%% The Appendices part is started with the command \appendix;
%% appendix sections are then done as normal sections
%% \appendix

%% \section{}
%% \label{}

\newpage

\appendix
\section{Randomized approximation of polynomial kernels (polynomial sketches)}
\label{app:polynomial-sketches}
% Authors including an appendix section should do so before References section. Multiple appendices should all have headings in the style used above. They will automatically be ordered A, B, C etc.

%\subsection{Example of a sub-heading within an appendix}
% There is also the option to include a subheading within the Appendix if you wish.

We define a random feature map $\phi_n$, from now on called a \textit{polynomial sketch}, as:
% maybe rather matrix form!
\begin{equation}
    \label{eqn:polynomial-estimator}
    \phi_n(\mat{x})
    :=  (\mat{W}_1 \mat{x} \odot \cdots \odot \mat{W}_n \mat{x}) / \sqrt{D} \in \mathbb{R}^D,
    \quad \text{($\odot$ denotes the element-wise product)}
\end{equation}
where $\mat{W}_1, \dots, \mat{W}_n \in \mathbb{R}^{D \times d}$ are %independently sampled
i.i.d. 
random matrices. For two inputs $\mat{x}, \mat{y} \in \mathbb{R}^d$ we have $\mathbb{E} [\phi_n(\mat{x})^{\top} \phi_n(\mat{y})] = (\dotprod{x}{y})^n$, i.e., the approximation is unbiased, if the $\{\mat{W}_i\}_{i=1}^n$ are sampled from an appropriate distribution.

Table~(\ref{tbl:sampling-variances}) shows three example sampling procedures along with the resulting variances $\mathbb{V}[\phi_n(\mat{x})^{\top} \phi_n(\mat{y})]$ of the corresponding kernel estimate. As shown in \cite{wacker2022a}, the variance of the Rademacher estimator is upper-bounded by the variance of the Gaussian estimator. TensorSRHT is a structured polynomial sketch that imposes an orthogonality constraint on the rows of each $\mat{W}_i$ leading to even lower variances for odd degrees $n$. We summarize its construction in Algorithm~(\ref{alg:tensor-srht-algorithm}), where the weights $\mat{W}_i$ are \textit{implicitly} defined. It uses the Fast Walsh-Hadamard Transform (FWHT) \cite{Fino1976} to project a single datapoint in $\bigO(n (D \log d))$ instead of $\bigO(n D d)$ time required for Gaussian and Rademacher sketches.

In this work, we only make use of Rademacher and TensorSRHT sketches as they yield the lowest variances. It is also possible to use complex-valued random matrices in Eq. (\ref{eqn:polynomial-estimator}) that yield additional variance reductions for positively valued data \cite{wacker2022b}. However, this condition does not hold for the method proposed in this work, which is why we consider only real-valued polynomial sketches here.

\begin{table}[t]
\caption{Polynomial Sketches according to \cite{wacker2022a} along with the variances of the associated kernel estimate. Here, $D \in \mathbb{N}$ is the number of random features
and $c(D, d) := \lfloor D/d \rfloor d(d-1) + (D \mod d)(D \mod d - 1)$.}
\label{tbl:sampling-variances}
\resizebox{\linewidth}{!}{%
\begin{tabular}{lll}
\toprule
Polynomial Sketch & Sampling Procedure for $\{\mat{W}_i\}_{i=1}^n$ & Variance $\mathbb{V}[\phi_n(\mat{x})^{\top} \phi_n(\mat{y})]$ \\
\midrule
Gaussian &
Entries of $\mat{W}_i$ are sampled i.i.d. from $\mathcal{N}(0, 1)$ &
$D^{-1} [ ( \norm{x}^2 \norm{y}^2 + 2 (\dotprod{x}{y})^2 )^n - (\dotprod{x}{y})^{2n} ]$ \\
Rademacher &
Entries of $\mat{W}_i$ are sampled i.i.d. from $\mathrm{Unif}(\{1, -1\})$ &
$D^{-1} [ ( \norm{x}^2 \norm{y}^2 + 2 ((\dotprod{x}{y})^2 - \sum_{i=1}^d x_i^2 y_i^2))^n - (\dotprod{x}{y})^{2n} ]$
\\
TensorSRHT &
$\mat{W}_i$ is implicitly defined through Algorithm (\ref{alg:tensor-srht-algorithm}) &
Rademacher variance $- \frac{c(D,d)}{D^2} \cdot$ \\
& &
$\Big[ (\mat{x}^\top \mat{y})^{2n} - \Big( ( \mat{x}^\top \mat{y} )^2 - \frac{1}{d-1} \Big( \|\mat{x}\|^2 \|\mat{y}\|^2 + ( \mat{x}^\top \mat{y} )^2 - 2 \sum_{k=1}^d x_k^2 y_k^2 \Big) \Big)^n \Big]$ \\
\bottomrule
\end{tabular}
}
\end{table}

\begin{algorithm}[t]
\caption{TensorSRHT according to \cite{wacker2022a}}
\label{alg:tensor-srht-algorithm}
    \SetAlgoLined
    \KwResult{A feature map $\phi_n(\mat{x})$}
    %initialization\;
    Pad $\mat{x}$ with zeros so that $d$ becomes a power of $2$ and let $\mat{H}_d \in \{1, -1\}^{d \times d}$ be the unnormalized Walsh-Hadamard matrix \cite{Fino1976};
    Let $B = \left\lceil \frac{D}{d} \right\rceil$ be the number of stacked projection blocks \;

    \ForAll{$b \in \{1, \dots, B\}$}{ 
        \ForAll{$i \in \{1, \dots, n\}$}{
            Generate a random vector ${\bf d}_i = (d_{i,1}, \dots, d_{i,d})^\top \in  \mathbb{R}^d$ as $d_{i,1}, \dots, d_{i,d} \stackrel{i.i.d.}{\sim} {\rm Unif}( \{ 1, - 1\} )$;
            % which is recursively defined through \\
            % $\mat{H}_{2n} :=
            % \begin{bmatrix}
            %     \mat{H}_n & \mat{H}_n \\
            %     \mat{H}_n & -\mat{H}_n
            % \end{bmatrix}$
            % and
            % $\mat{H}_2 :=
            % \begin{bmatrix}
            %     1 & 1\\
            %     1 & -1
            % \end{bmatrix}$;
            
            Compute $\phi^{b,i}(\mat{x}) := \mat{H}_d (\mat{d}_i \odot \mat{x})$ using the FWHT \cite{Fino1976} and shuffle the elements of $\phi^{b,i}(\mat{x})$ randomly; % define H
        }
        Compute $\phi^b(\mat{x}) := (\phi^{b,1}(\mat{x}) \odot \cdots \odot \phi^{b,n}(\mat{x})) / \sqrt{D}$ \;
    }
  
    Concatenate the elements of $\phi^1(\mat{x}), \dots, \phi^B(\mat{x})$ to yield a single vector $\phi_n(\mat{x})$ and keep the first $D$ entries \;
    
    %\textit{Note: This is akin to partitioning the rows of each $\{\mat{W}_i\}_{i=1}^n$ in Eq.~(\ref{eqn:polynomial-estimator}) into $B$ blocks. Each block is an independently sampled version of $\mat{H}_d {\rm diag}(\mat{d}_i)$ with shuffled rows. ${\rm diag}(\mat{d}_i)$ is a matrix with $\mat{d}_i$ on the diagonal.}
\end{algorithm}

\section{Proof of Theorem \ref{thrm:norm-decay}}
\label{app:thrm-proof}

\begin{proof}
We start by deriving an upper bound for $|k_p(\mat{x}, \mat{y})|$. We leave out the length scale here for ease of notation.
\begin{align*}
    |k_p(\mat{x}, \mat{y})|
    = \sigma^2 \exp(- (\|\mat{x}\|^2 + \|\mat{y}\|^2) / 2) \left| \sum_{n=0}^p 1/n! (\dotprod{x}{y})^n \right|
    &\leq \sigma^2 \exp(- (\|\mat{x}\|^2 + \|\mat{y}\|^2) / 2) \sum_{n=0}^p 1/n! (\norm{x} \norm{y})^n \\
    %&\leq \sigma^2 \exp(- (\|\mat{x}\|^2 + \|\mat{y}\|^2) / 2) \sum_{n=0}^p 1/n! ((\norm{x}^2 + \norm{y}^2) / 2)^n \\
    %&\leq \sigma^2 \exp(- (\|\mat{x}\|^2 + \|\mat{y}\|^2) / 2) \exp((\norm{x}^2 + \norm{y}^2) / 2)
    %= \sigma^2
    %&\leq \sigma^2 \exp(- (\|\mat{x}\|^2 + \|\mat{y}\|^2) / 2) \exp ((\norm{x}^2 + \norm{y}^2) / 2) = \sigma^2
\end{align*}
Next, we notice that $\|\mat{x} - \mat{y}\|^2 = \norm{x}^2 + \norm{y}^2 - 2 \dotprod{x}{y} \geq 0$ for any $\mat{x}, \mat{y}$. Thus, we can choose them to be parallel. So $\norm{x} \norm{y} \leq (\norm{x}^2 + \norm{y}^2) / 2$. From this inequality, it follows
\begin{align*}
    \sum_{n=0}^p 1/n! (\norm{x} \norm{y})^n
    \leq \sum_{n=0}^p 1/n! ((\norm{x}^2 + \norm{y}^2) / 2)^n
    \leq \exp((\|\mat{x}\|^2 + \|\mat{y}\|^2) / 2).
\end{align*}
Now, the gap $\exp((\|\mat{x}\|^2 + \|\mat{y}\|^2) / 2) - \sum_{n=0}^p 1/n! ((\|\mat{x}\|^2 + \|\mat{y}\|^2) / 2)^n = \sum_{n=p+1}^\infty 1/n! ((\|\mat{x}\|^2 + \|\mat{y}\|^2) / 2)^n$ increases as $\norm{x}^2 + \norm{y}^2$ increases, which must decreases the ratio $\sum_{n=0}^p 1/n! (\norm{x} \norm{y})^n / \exp((\norm{x}^2 + \norm{y}^2) / 2)$ and thus the upper bound of $|k_p(\mat{x}, \mat{y})|$ goes to zero as $\norm{x}^2 + \norm{y}^2$ increases.

Next, we look at the error $|k(\mat{x}, \mat{y}) - k_p(\mat{x}, \mat{y})|
= \sigma^2 \exp(- (\|\mat{x}\|^2 + \|\mat{y}\|^2) / 2) \left| \sum_{n=p+1}^\infty 1/n! (\dotprod{x}{y})^n\right|$.
If the angle between $\mat{x}$ and $\mat{y}$ is zero such that $\dotprod{x}{y} = \norm{x} \norm{y}$, all addends in the infinite sum are maximized. The error thus becomes the largest. The error is zero when they are orthogonal.
\end{proof}

%% References
%%
%% Following citation commands can be used in the body text:
%% Usage of \cite is as follows:
%%   \cite{key}         ==>>  [#]
%%   \cite[chap. 2]{key} ==>> [#, chap. 2]
%%

%The citation must be used in following style: \cite{article-minimal} \cite{article-full} \cite{article-crossref} \cite{whole-journal}.
%% References with BibTeX database:

\printbibliography
%\bibliographystyle{elsarticle-harv}

%% Authors are advised to use a BibTeX database file for their reference list.
%% The provided style file elsarticle-num.bst formats references in the required Procedia style

%% For references without a BibTeX database:

% \begin{thebibliography}{}

%% \bibitem must have the following form:
%%   \bibitem{key}...
%%

% \bibitem{Massimo2011}
% {F}ilippini, Massimo, and Lester C. Hunt. (2011) ``Energy demand and
% energy efficiency in the OECD countries: a stochastic demand frontier
% approach." {\it Energy Journal} {\bf 32} (2): 59--80.
% \bibitem{Massimo2012}
% Filippini, Massimo, and Lester C. Hunt. (2012) ``US residential
% energy demand and energy efficiency: A stochastic demand frontier
% approach." {\it Energy Economics} {\bf 34} (5): 1484--1491.
% \bibitem{Thomas2015} 
% Weyman-Jones, Thomas, J\'{u}lia Mendon\c{c}a Boucinha, and Catarina
% Feteira In\'{a}cio. (2015) ``Measuring electric energy efficiency in
% Portuguese households: a tool for energy policy." {\it Management of Environmental Quality: An International Journal} {\bf 26} (3): 407--422.
% \bibitem{} 
% Saunders, Harry (2009) ``Theoretical Foundations of the Rebound Effect'', in Joanne Evans and Lester Hunt (eds) {\it International Handbook on the Economics of Energy}, Cheltenham, Edward Elgar
% \bibitem{} 
% Sorrell, Steve (2009) ``The Rebound Effect: definition and estimation'', in Joanne Evans and Lester Hunt (eds) {\it International Handbook on the Economics of Energy}, Cheltenham, Edward Elgar 
% \end{thebibliography}

\end{document}